\documentclass{IEEEcsmag}

\usepackage[colorlinks,urlcolor=blue,linkcolor=blue,citecolor=blue]{hyperref}
\expandafter\def\expandafter\UrlBreaks\expandafter{\UrlBreaks\do\/\do\*\do\-\do\~\do\'\do\"\do\-}
\usepackage{upmath,color}
\usepackage{amsmath,amssymb,amsfonts,makecell,algorithm,algorithmic,multirow} 

\begin{document}


\title{Feature Importance Guided Random Forest Learning with Simulated Annealing Based Hyperparameter Tuning}

\author{Kowshik Balasubramanian}
\affil{Florida Atlantic University, Boca Raton, FL, 33431 USA}

\author{Andre Williams}
\affil{Florida Atlantic University, Boca Raton, FL, 33431 USA}

\author{Ismail Butun}
\affil{Florida Atlantic University, Boca Raton, FL, 33431 USA}


\begin{abstract}\looseness-1This paper introduces a novel framework for enhancing Random Forest classifiers by integrating probabilistic feature sampling and hyperparameter tuning via Simulated Annealing. The proposed framework exhibits substantial advancements in predictive accuracy and generalization, adeptly tackling the multifaceted challenges of robust classification across diverse domains, including credit risk evaluation, anomaly detection in IoT ecosystems, early-stage medical diagnostics, and high-dimensional biological data analysis. To overcome the limitations of conventional Random Forests, we present an approach that places stronger emphasis on capturing the most relevant signals from data while enabling adaptive hyperparameter configuration. The model is guided towards features that contribute more meaningfully to classification and optimizing this with dynamic parameter tuning. The results demonstrate consistent accuracy improvements and meaningful insights into feature relevance, showcasing the efficacy of combining importance aware sampling and metaheuristic optimization.
\end{abstract}

\maketitle

\chapteri{T}he increasing complexity of modern data in finance, industrial Internet Of Things (IoT), healthcare, and bioinformatics demands machine learning models that are both interpretable and high-performing. Conventional Random Forests (RF), while powerful, often struggle with feature redundancy, parameter sensitivity, and domain-specific biases. Our Feature Importance Guided Random Forest (FIGRF) framework addresses these challenges through feature-importance-guided sampling and Simulated Annealing (SA) optimization, enabling consistently superior results across anomaly detection, biomedical diagnostics, and standard machine learning benchmarks.

RFs are widely used ensemble learning methods known for their robustness, interpretability, scalability and performance across diverse machine learning tasks. However, the standard approach to feature sampling in RF typically involves uniform random selection, which can underutilize important features and introduce noise into the model. Furthermore, hyperparameter optimization is often handled through grid or random search, both of which are computationally expensive.

To address these limitations, we propose a two-fold enhancement to the classical RF framework. First, we develop a probabilistic feature sampling strategy that emphasizes informative features by computing and combining three widely recognized importance measurement techniques - Permutation Importance, Gini Importance, and Mutual Information. These importance scores are normalized, averaged, and transformed into a sampling probability distribution via softmax, enabling the model to prioritize features with high relevance and minimize the use of features with low relevance in building RF while retaining ensemble diversity of feature selection techniques.

Second, we employ SA - a metaheuristic stochastic global optimization technique inspired by the physical annealing process to efficiently search for optimal hyperparameters. By adopting a temperature controlled exploration-exploitation trade-off, this method escapes local optima and converges to high-performing configurations for the number of estimators and tree depth.

The proposed methodology not only improves classification performance but also provides feature usage insights and generalizes well across datasets. We validate our approach on seven publicly available binary classification datasets and compare it against standard RF baseline model to demonstrate consistent improvements in accuracy, precision, recall, F1-score and advancing modern internet systems. The code is available in \url{https://github.com/KowshikB03/Feature-Importance-Guided-Random-Forest}

Recent advancements in optimizing Random Forest feature utilization include hybrid approaches. One method combines the Seagull Optimization Algorithm for feature selection with Random Forest classification, improving robustness by removing redundant features and leveraging RF’s ability to handle high-dimensional noisy data \cite{yaqoobrandomforest}. Another stduy employs a layered process where RF importance scores perform preliminary elimination, followed by a genetic algorithm to refine the subset, enhancing accuracy and search efficiency in large feature spaces \cite{dingrandomforest}.

\section{2. Related Works}
\label{2}

Our research integrates concepts from several prior works, and we compare them against our approach. Similar to Nguyen et al. \cite{nguyen2015unbiased} and Liu \& Zhao \cite{liu2017variable}, the proposed Feature Importance Guided Random Forest (FIGRF) framework uses feature importance scores to guide Random Forest construction, but instead of relying on a single metric or re-weighted sampling, it adopts a composite multi-metric strategy with probabilistic softmax sampling for more flexibility. Compared to the Canonical Correlation Analysis–based method in \cite{liu2018exploring}, which iteratively selects features through linear correlations, our ensemble method captures nonlinear dependencies and complex interactions more effectively. Altmann et al. \cite{altmann2010permutation} proposed a statistically corrected permutation importance approach with significance testing, while our implementation averages accuracy drops across multiple shuffles. Reinforcement-learning-based feature selection \cite{bouchlaghem2024novel} and Chi2-PSOGWO \cite{abdo2024optimized} provide adaptive and hybrid strategies but are computationally expensive and prone to redundancy or repeated evaluations. The mutual information–based framework in \cite{peng2005feature}, particularly the mRMR criterion, inspired our mutual information component, which is combined with permutation and Gini importance to mitigate individual biases noted in \cite{altmann2010permutation, bouchlaghem2024novel, abdo2024optimized, peng2005feature}. Finally, various studies have extended SA, such as hybridization with Atom Search Optimization \cite{ghosh2020aso}, pressure sensor optimization in water networks \cite{morales2021pressure}, and multi-objective CNN hyperparameter tuning \cite{gulcu2021multi}. In a similar way, we employ SA for efficient hyperparameter optimization of the FIGRF model trained with probabilistic feature selection.

\section{3. Feature Importance Selection Strategies}
\label{3}
\vspace{0pt}
Permutation Importance, Gini Importance and Mutual Information approaches provide a diverse and comprehensive evaluation of feature significance. However, when considered in isolation, each method is prone to certain limitations including sensitivity to feature correlation, biases toward features with specific value distributions or cardinalities and lack of model context in purely statistical measures can lead to inconsistent or misleading feature contributions. Since each method is effective for different dataset structures, we combined them into a composite framework that normalizes and averages their importance scores. Then, applying softmax-adjusted normalization further offsets individual biases and stabilizes the estimates, enhancing the robustness and reliability of feature selection.

\subsection{3.1 Permutation Importance}
\label{3.1}

Permutation Importance is a model-agnostic technique used to evaluate the contribution of each feature by quantifying the impact of randomly shuffling its values on model performance. As discussed in \cite{fisher2019all}, permutation-based importance is defined by disrupting the relationship between a feature and the target, and measuring the corresponding drop in predictive performance to estimate that feature's relevance. Following this principle, our implementation begins by training a model on the original dataset and recording its baseline accuracy, \( Acc_{\text{base}} \), on a validation set. For a given feature \( f \), we then permute its values across all samples, breaking its association with the target variable, and evaluate the model on this perturbed data to obtain \( Acc_{\text{perm}}^{(r)}(f) \).

To mitigate randomness introduced by a single shuffle, this process is repeated \( R \) times, and the feature’s importance is calculated as the mean drop in accuracy across these trials. A larger drop
implies the feature is more important, since its disruption leads to a significant degradation in the model’s predictive performance. The final importance score is computed as:

\begin{equation}
P(f) = \frac{1}{R} \sum_{r=1}^{R} \left( Acc_{\text{base}} - Acc_{\text{perm}}^{(r)}(f) \right)
\tag{1}
\label{eq:perm}
\end{equation}

Equation ~\eqref{eq:perm}, \( P(f) \) denotes the permutation importance score for feature \( f \), \( Acc_{\text{base}} \) is the original model accuracy, and \( Acc_{\text{perm}}^{(r)}(f) \) is the accuracy after the \( r \)-th permutation of feature \( f \). Averaging across multiple random permutations ensures a more stable and robust estimate of feature importance.

\subsection{3.2 Gini Importance}
\label{3.2}

In a decision tree, each split is selected in a way to maximize the reduction in impurity. Gini impurity is a measure of how often a randomly chosen element would be incorrectly labeled if it is randomly labeled according to the distribution of labels in the node. When a feature is used to split a node, it partitions the data to achieve more homogeneous subgroups, thus reducing impurity. It quantifies the importance of a feature based on how frequently and effectively it is used to split the data across all the decision nodes in the model. The greater the reduction in impurity, the more informative the feature is considered to be for classification.

The Gini Importance for a feature \( f \) is calculated by summing the impurity reductions \( \Delta \text{Gini}(n) \) over all the nodes \( n \) in which the feature is used, across all trees \( T \). Each reduction is weighted by the proportion \( p(n) \) of samples reaching that node, ensuring that splits affecting more data points contribute more to the overall importance.

\begin{equation}
G(f) = \sum_{t=1}^{T} \sum_{n \in N_t(f)} p(n) \cdot \Delta \text{Gini}(n)
\tag{2}
\label{eq:gini}
\end{equation}

A higher Gini Importance score indicates that the feature plays a more critical role in improving the purity of the nodes and therefore has a greater influence on the model's decisions. Our use of Gini importance is motivated by the analysis in \cite{louppe2013understanding}, which provides theoretical insights into how this measure behaves in randomized trees and highlights its potential biases. Their findings guided our decision to normalize and combine Gini importance with other measures from Sections \href{3.1}{3.1} and \href{3.3}{3.3} to obtain a more balanced view of feature relevance.

\subsection{3.3 Mutual Information}
\label{3.3}
As discussed in Section \hyperref[2]{2}, \cite{peng2005feature} proposes a mutual information-based feature selection framework that introduced the concepts of max-dependency, max-relevance, and min-redundancy to select features that are informative with respect to the target while minimizing inter-feature redundancy. Building on their max-relevance criterion, we use Mutual Information (MI) to quantify the dependency between each feature and the target variable by measuring the reduction in uncertainty of one variable given knowledge of the other.

The computation is based on the concept of entropy and involves estimating the joint probability distribution \( P(x_f, y) \) along with the marginal distributions \( P(x_f) \) and \( P(y) \). Specifically, for each pair of discrete values \( (x_f, y) \), the contribution to MI is calculated as the product of the joint probability and the logarithm of the ratio between the joint and the product of marginal probabilities. This process requires discretizing continuous features (if present), constructing contingency tables or histograms, and normalizing counts to obtain empirical probability distributions.

\begin{equation}
MI(f; Y) = \sum_{x_f} \sum_{y} P(x_f, y) \log \left( \frac{P(x_f, y)}{P(x_f)P(y)} \right)
\tag{3}
\label{eq:mi}
\end{equation}

In equation ~\eqref{eq:mi}, the logarithmic term evaluates the divergence from independence; when the joint probability equals the product of the marginals, the term becomes zero, indicating no mutual information. In practice, the accuracy of MI estimation depends heavily on the granularity of discretization and the sample size, as poor estimation of probabilities can lead to biased results.

While \cite{peng2005feature} includes redundancy minimization (mRMR) via pairwise MI between features, we intentionally exclude this step. Instead, we use MI solely to assess feature-to-target relevance and combine the normalized MI scores with Gini and permutation importance to form a composite metric that guides feature sampling probabilities in our feature selection ensemble.

\subsection{3.4 Averaging and Softmax Normalization}

To effectively combine the three feature importance scores from sections \hyperref[3.1]{3.1}, \hyperref[3.2]{3.2} and \hyperref[3.3]{3.3}, we apply a two-step normalization and transformation strategy. This method is used to ensure scale consistency and probabilistic interpretation, facilitating stochastic feature selection.

First, each importance score is normalized using min-max scaling to bring the values into a common range \([0, 1]\). This prevents features with inherently higher raw scores (like the impurity values formulated in Section \hyperref[3.2]{3.2}) from dominating the combined ranking. During implementation, each importance scores is independently scaled across all features. Then, we compute the average normalized importance score for each feature across all trees in the ensemble model.
\begin{equation}
\bar{f_i} = \frac{1}{T} \sum_{t=1}^{T} f_{i}^{(t)}
\tag{4}
\label{eq:norm}
\end{equation}

where \( f_{i}^{(t)} \) denotes the normalized importance of feature \( i \) in tree \( t \), and \( T \) is the total number of trees in equation ~\eqref{eq:norm}. This aggregation is implemented as a mean operation across all trees using vectorized operations. Next, features are sorted in descending order according to their average importance scores to rank all features importance.
\begin{equation}
\bar{f}_{(1)} \geq \bar{f}_{(2)} \geq \dots \geq \bar{f}_{(d)}
\tag{5}
\label{eq:desc}
\end{equation}

where \( d \) is the total number of features and \( \bar{f}_{(d)} \) represents the \( d \)-th highest average score in equation ~\eqref{eq:desc}. Sorting is applied once the average importance vector is obtained. Finally, we convert the sorted average importance scores into a discrete probability distribution using the softmax function in equation~\eqref{eq:softmax}.
\begin{equation}
p_i = \frac{e^{\alpha \cdot \bar{f}_i}}{\sum_{j=1}^{d} e^{\alpha \cdot \bar{f}_j}}, \quad \text{for } i = 1, \dots, d
\tag{6}
\label{eq:softmax}
\end{equation}
where:
\begin{itemize}
    \item \( \bar{f}_i \) is the min-max normalized average importance score of feature \( i \)
    \item \( \alpha \) is the temperature scaling parameter, controlling the distribution sharpness

\end{itemize}

The softmax transformation converts importance scores into sampling probabilities, enabling stochastic feature selection based on relative importance. In practice, higher \( \alpha \) values result in more peaked distributions, emphasizing top-ranked features, while lower values yield a more uniform spread. In this case, the selection of \( \alpha \) is performed by trial-and-error, based on the structure of each dataset.\vspace*{-5pt}

\section{4. Feature Importance Guided Random Forest}
\label{4}

\subsection{4.1 Probabilistic Feature Sampling}

Traditional RF selects a random subset of features uniformly at each tree. In contrast, our approach performs \textit{probabilistic feature sampling}, where features are selected based on the probability distribution \( \mathbf{p} = [p_1, p_2, \dots, p_d] \) obtained from Section \hyperref[3]{3}. For each tree in the forest, instead of uniform random sampling of features, we sample from \( \mathbf{p} \) favoring more informative variables. The training procedure for each tree \( h^{(t)} \), \( t = 1, \dots, T \), is as follows:

\begin{enumerate}
    \item Draw a bootstrap sample of the training data by randomly selecting \( n \) rows with replacement.
    \item Select \( m = \lfloor \sqrt{d} \rfloor \) features using weighted sampling \textbf{without replacement} from the distribution \( \mathbf{p} \).
    \item Train a decision tree \( h^{(t)} \) using only the selected \( m \) features on the sampled data.
\end{enumerate}

The final prediction \( \hat{y} \) for an input sample \( \mathbf{x} \in \mathbb{R}^d \) is obtained via majority voting:
\[
\hat{y} = \text{majority\_vote} \left( \left\{ h^{(t)}(\mathbf{x}_{S^{(t)}}) \right\}_{t=1}^{T} \right)
\]
where:
\begin{itemize}
    \item \( S^{(t)} \subset \{1, \dots, d\} \) is the index set of features selected for tree \( t \) based on \( \mathbf{p} \),
    \item \( \mathbf{x}_{S^{(t)}} \) is the projection of the input vector \( \mathbf{x} \) onto the selected features,
    \item \( T \) is the total number of trees in the ensemble.
\end{itemize}

During inference, each tree predicts using only the features it is trained on. The predictions across all trees are averaged (or majority voted), and the final class label is returned, maintaining the ensemble diversity of standard RF. The mechanism prioritizes informative features while maintaining diversity through probabilistic exploration, balancing bias and variance. It improves efficiency by reducing focus on uninformative features, accelerates training by lowering the number of trees needed, and mitigates overfitting by ensuring feature sampling preserves ensemble diversity.

\subsection*{4.2 Feature Usage Tracking}
\label{4.2}
The framework logs how frequently each feature is used across trees \( T \) in the RF in percentage, displaying insights into feature utilization in the final FIGRF Model:
\[
Usage(f) = \sum_{t=1}^{T} \mathbb{1}\{f \in Features_t\}
\]

\section{5. Hyperparameter Tuning via Simulated Annealing}
\label{5}

\begin{figure*}[h]

\centerline{\includegraphics[width=30pc]{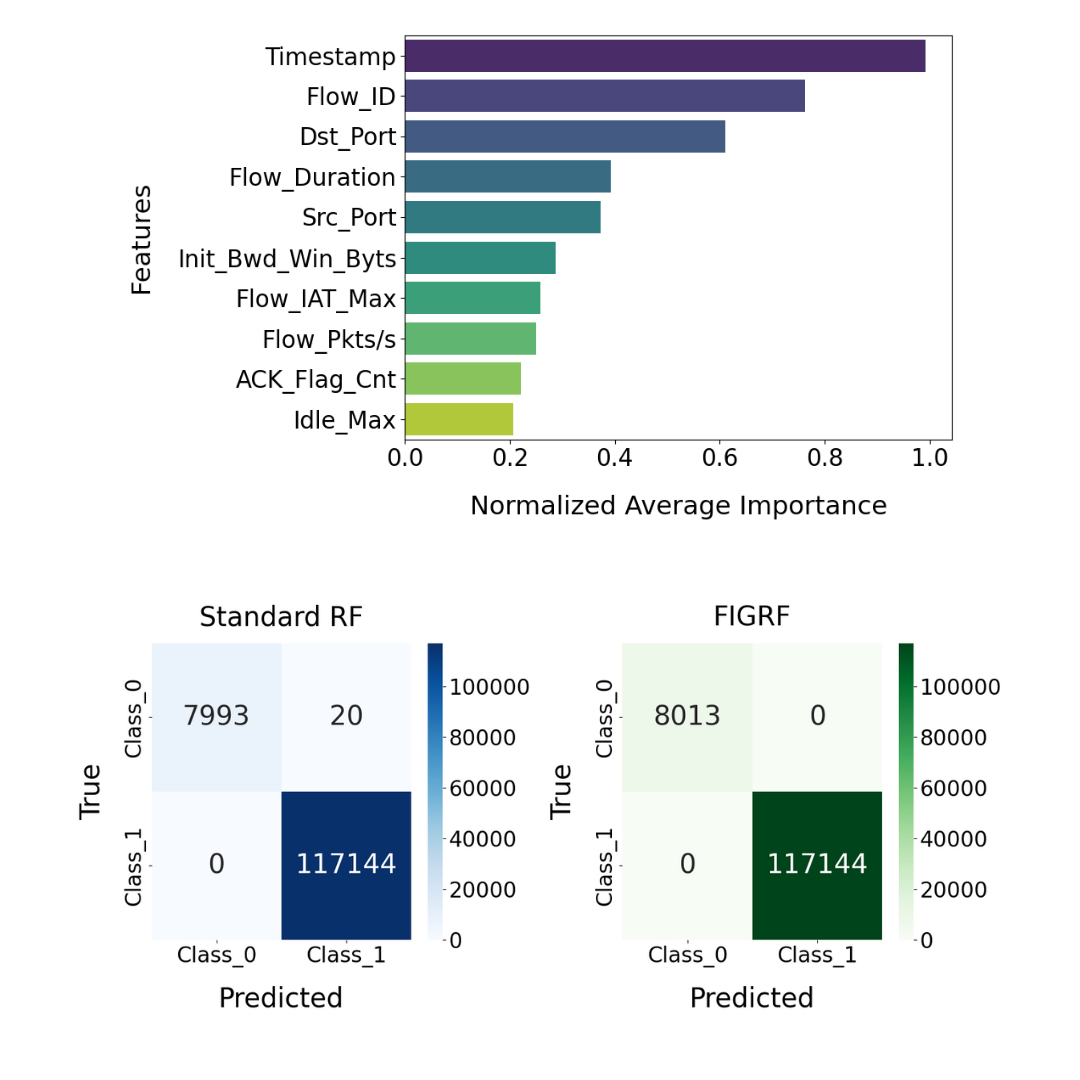}}
    \caption{Top ten important features and Standard vs FIGRF performance - IOTID20 Dataset}
    \label{imp_iotid20}\vspace*{-5pt}
\end{figure*}

Hyperparameter tuning is essential to prevent overfitting and improve the performance of our RF model, since the ensemble feature selection in Section \hyperref[4]{4} may introduce bias in certain cases. To address this, we adopt Simulated Annealing (SA) to optimize two key hyperparameters: the number of estimators (\(n_{estimators}\)) and maximum depth (\(max\_depth\)) of each decision tree. As analyzed by Ingber \cite{ingber1993simulated}, SA offers practical advantages over grid or random search by efficiently exploring complex search spaces with fewer parameters, lower computational cost, and faster iterations than Genetic Algorithms, while its cooling schedule ensures gradual convergence. Our implementation follows the classical framework introduced by Kirkpatrick et al. \cite{kirkpatrick1983optimization}, which accepts worse solutions probabilistically to escape local optima and steadily reduces temperature to approach a global optimum. The optimization is directed toward maximizing a composite fitness function defined as the average of Accuracy, Precision, Recall, and F1-score, ensuring balanced improvements across classification metrics rather than favoring a single measure.

The SA algorithm starts with a randomly selected initial configuration within the predefined ranges:
\begin{itemize}
    \item \( n_{estimators} \) varies between 50 and 150,
    \item \( max\_depth \) is chosen from the set \(\{ 5, 6, 7, \ldots, 20\}\) or is left unconstrained (\(None\)).
\end{itemize}

The SA iteration starts with a randomly selected initial configuration. The number of estimators is chosen randomly between 50 and 150. For the maximum depth, the initial value is selected from a discrete set of integers from 5 to 20 or, alternatively, it is set to None, allowing the trees to grow to full depth without a size constraint. At each iteration, a small perturbation is applied to the current hyperparameter configuration to generate a neighboring candidate solution. This neighbor represents a slight modification in either \( n_{estimators} \) or \( max\_depth \). The fitness of the candidate solution is then evaluated by training the model with the candidate hyperparameters and measuring its performance on a validation set. 

To decide whether to move to the neighbor configuration, the algorithm uses a probabilistic acceptance criterion. If the neighbor yields an improved fitness (\(\Delta fitness > 0\)), it is always accepted. Otherwise, the neighbor can still be accepted with a probability that decreases exponentially with how much worse the neighbor is and the current "temperature" \( T \). This acceptance probability is given by:

\[
P_{accept} = 
\begin{cases}
1 & \text{if } \Delta fitness > 0 \\
\exp\left(\frac{\Delta fitness}{T}\right) & \text{if } \Delta fitness \leq 0
\end{cases}
\]

Here, \( \Delta fitness = fitness_{neighbor} - fitness_{current} \), and \( T \) controls the likelihood of accepting worse solutions. High temperatures allow more exploration by accepting worse solutions more frequently, helping the algorithm avoid local optima. As the temperature decreases over iterations, the algorithm becomes more conservative, focusing on exploitation around promising solutions.

The temperature is gradually reduced at each iteration according to the cooling schedule:

\[
T \leftarrow \alpha T
\]

where \( \alpha \in (0,1) \) is the cooling rate.

This iterative process continues until a stopping criterion is met, that is maximum number of iterations. The hyperparameter configuration with the best score found during the search is then selected for the final model training. When two hyperparameter configurations yield the same best score, it's prefered to select the one with a lower number of trees  \( n_{estimators} \) and a smaller tree depth \( max\_depth \) to prevent overfitting and improve generalization. This methodology provides a balance between exploration and exploitation in the hyperparameter search space, helping to find near-optimal parameters efficiently without exhaustive search.\vspace*{-5pt}

\section{6. Algorithm}
\hrulefill

\noindent \textbf{Algorithm 1:} Feature Importance Guided Random Forest Model with Simulated Annealing Optimization

\noindent \hrulefill

\begin{algorithmic}[1]
\STATE \textbf{Input:} Dataset $\mathcal{D} = \{(x_i, y_i)\}_{i=1}^N$, number of trees $T$, total features $d$
\STATE \textbf{Output:} Optimized hyperparameters $\theta^* = (n_{\text{estimators}}^*, \text{max\_depth}^*)$, trained model ensemble

\STATE \textbf{Step 1:} Compute feature importance scores

\STATE Compute Permutation Importance by ~\eqref{eq:perm} for each feature
\STATE Compute Gini Importance by ~\eqref{eq:gini} for each feature $f \in \{1, \ldots, d\}$
\STATE Compute Mutual Information by ~\eqref{eq:mi} for each feature

\STATE \textbf{Step 2:} Normalize and average importance scores
\FOR{each feature $f$}
    \STATE Normalize $P(f), G(f), \text{MI}(f)$ using min-max scaling
    \STATE Compute average importance $\bar{I}(f) = \frac{G'(f) + P'(f) + \text{MI}'(f)}{3}$
\ENDFOR

\STATE \textbf{Step 3:} Compute feature sampling probabilities
\STATE Sort features by descending $\bar{I}(f)$
\STATE Compute sampling probability distribution:
\[
p_f = \frac{\exp(\alpha \cdot \bar{I}(f))}{\sum_{j=1}^d \exp(\alpha \cdot \bar{I}(j))}
\]

\STATE \textbf{Step 4:} Train ensemble using probabilistic feature sampling
\FOR{tree $t = 1$ to $T$}
    \STATE Draw bootstrap sample from dataset with replacement
    \STATE Sample $m = \lfloor \sqrt{d} \rfloor$ features without replacement from $p_f$
    \STATE Train decision tree $h^{(t)}$ on sampled data using selected features
\ENDFOR

\STATE \textbf{Step 5:} Simulated Annealing Hyperparameter Tuning
\STATE Initialize temperature $\text{temp}$, cooling rate $\alpha$, initial configuration $\theta$
\STATE Initialize best configuration $\theta^* \leftarrow \theta$, best fitness $F^* \leftarrow -\infty$

\WHILE{stopping criterion not met}
    \STATE Generate neighbor $\theta'$ by small perturbation of $\theta$
    \STATE Evaluate fitness $F' = \text{Fitness}(\theta')$ on validation data
    \STATE Compute $\Delta F = F' - F$
    \IF{$\Delta F > 0$ or $\exp(\frac{\Delta F}{\text{temp}}) > \text{random}(0,1)$}
        \STATE Accept $\theta \leftarrow \theta'$, $F \leftarrow F'$
        \IF{$F' > F^*$ \OR ($F' = F^*$ \AND Complexity($\theta'$) < Complexity($\theta^*$)) }
            \STATE Update best configuration $\theta^* \leftarrow \theta'$, best fitness $F^* \leftarrow F'$
        \ENDIF

    \ENDIF
    \STATE Decrease temperature $\text{temp} \leftarrow \alpha \times \text{temp}$
\ENDWHILE

\STATE \textbf{Return} best hyperparameters $\theta^*$
\end{algorithmic}
\noindent \hrulefill

\section{7. Experimental Setup}
\vspace{3pt}

\begin{table*}[htbp]
\centering
\vspace*{4pt} 
\caption{Standard vs FIGRF Performance}
\label{tab:rf_performance_comparison}
\begin{tabular}{|l|l|l|}
\hline
\textbf{Dataset} & 
\makecell{\textbf{Standard RF Performance}\\(Accuracy, Precision, Recall, F1)} & 
\makecell{\textbf{FIGRF Performance}\\(Accuracy, Precision, Recall, F1)} \\ \hline

Credit Card & 
81.53\%, 63.61\%, 36.48\%, 46.37\% & 
82.1\%, 70\%, 31.76\%, 43.71\% \\ \hline

IOTID20 & 
99.98\%, 99.98\%, 100\%, 99.99\% & 
100\%, 100\%, 100\%, 100\% \\ \hline

Bank Marketing & 
91.24\%, 65.24\%, 48.77\%, 55.81\% & 
91.54\%, 64.8\%, 55.72\%, 59.91\% \\ \hline

WUSTL IIOT 2021 & 
99.99\%, 99.99\%, 99.99\%, 99.99\% & 
99.99\%, 99.99\%, 99.99\%, 99.99\% \\ \hline

Darwin & 
94.29\%, 94.29\%, 94.29\%, 94.29\% & 
100\%, 100\%, 100\%, 100\% \\ \hline

Iris & 
100\%, 100\%, 100\%, 100\% & 
100\%, 100\%, 100\%, 100\% \\ \hline

Wine & 
100\%, 100\%, 100\%, 100\% & 
100\%, 100\%, 100\%, 100\% \\ \hline
\end{tabular}\vspace*{8pt} 
\end{table*}

\begin{table*}[htbp]
\centering
\vspace*{4pt} 
\caption{Related Works vs FIGRF Performance}
\label{tab:rf_performance_comparison_RW}
\begin{tabular}{|c|c|c|}
\hline
\textbf{Dataset} & 

\makecell{\textbf{Related Works Performance}\\(Accuracy)} &
\makecell{\textbf{FIGRF Performance}\\(Accuracy)} \\ \hline

Credit Card & 
FS-RL \cite{bouchlaghem2024novel}: 78.9\% & 
82.1\% \\ \hline

Iris & 
Chi2-PSOGWO \cite{abdo2024optimized}: 95\% & 
100\% \\ \hline

\multirow{2}{*}{Wine} & 
Chi2-PSOGWO \cite{abdo2024optimized}: 99.4\% & 
100\% \\ \cline{2-3}
 & ASOs+SA \cite{ghosh2020aso}: 100\% & 
100\% \\ \hline
\end{tabular}\vspace*{8pt} 
\end{table*}

\begin{table*}[htbp]
\centering
\vspace*{4pt} 
\caption{Simulated Annealing tuned hyperparameters of FIGRF}
\label{tab:rf_hyperparameters}
\begin{tabular}{|l|c|c|}
\hline
\textbf{Dataset} & \textbf{n\_estimators} & \textbf{max\_depth} \\ \hline

Credit Card & 140 & 20 \\ \hline
IOTID20 & 100 & 18 \\ \hline
Bank Marketing & 50 & 8 \\ \hline
WUSTL IIOT 2021 & 60 & 20 \\ \hline
Darwin & 53 & 10 \\ \hline
Iris & 71 & 10 \\ \hline
Wine & 109 & 13 \\ \hline

\end{tabular}\vspace*{8pt} 
\end{table*}

\begin{figure*}[h]
\centerline{\includegraphics[width=30pc]{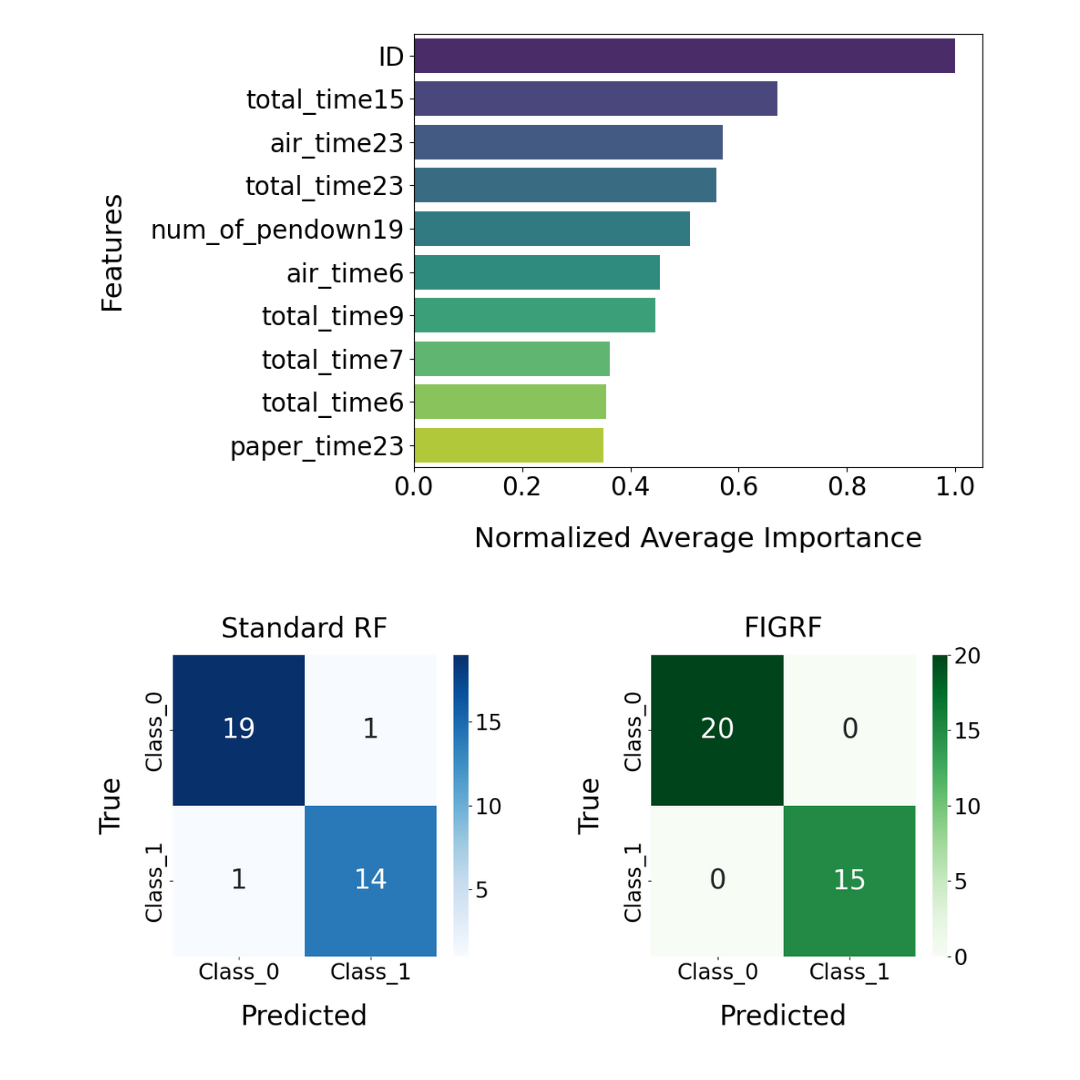}}
    \caption{Top ten important features and Standard vs FIGRF performance - UCI Darwin Dataset}
    \label{imp_darwin}
\end{figure*}

The FIGRF framework was evaluated on seven benchmark datasets using the setup described in Section \hyperref[5]{5}, with experiments implemented in Scikit Learn \cite{scikit-learn}. Preprocessing steps varied by dataset but generally included label encoding categorical variables, imputing missing or outlier values, and standardizing numerical features using \texttt{StandardScaler}. A baseline \texttt{RandomForestClassifier} with 100 estimators and default parameters was used for comparison. Feature importances were computed via 2-fold permutation importance, Gini importance from the trained RF, and mutual information (\texttt{random\_state=42}). These scores were normalized, averaged, and transformed into feature sampling probabilities using softmax scaling. The FIGRF model was then trained on these probabilities, with the number of trees and tree depth optimized using SA as described in Section \hyperref[5]{5}. SA was run for 30 iterations in most cases, which provided convergence to optimal configurations, except for the WUSTL-IIoT 2021 dataset, which required additional iterations. To ensure unbiased results, train–test splits (80:20) were performed before preprocessing, scaling was fit only on training data, and feature importance along with hyperparameter tuning were calculated exclusively on training data; the test set remained unseen until final evaluation. Model performance was assessed using accuracy, precision, recall, F1-score, and confusion matrices compared against the baseline. The datasets included both high-dimensional and low-dimensional cases. The UCI Credit Card Client dataset (30,000 samples) was preprocessed by removing the ID column, renaming the target variable, and applying one-hot encoding to categorical features such as \texttt{SEX}, with $\alpha = 5$ for softmax scaling. The IoTID20 dataset \cite{ullah2020scheme} was converted from multi-class to binary classification by removing categorical columns, imputing missing values with a placeholder string, encoding categorical features with \texttt{LabelEncoder}, and mapping ``Normal'' to 0 and ``Anamoly'' to 1, with $\alpha = 5$. The UCI Bank Marketing dataset (41,188 samples) was processed by applying one-hot encoding with the first category dropped, defining the binary target as \texttt{y\_yes}, and setting $\alpha = 5$. The WUSTL-IIoT 2021 dataset \cite{zolanvari2021wustl} was tailored for IIoT anomaly detection, with columns such as \texttt{Traffic}, \texttt{StartTime}, \texttt{LastTime}, and IP addresses removed to avoid target leakage, categorical features imputed with ``missing,'' numeric features imputed with medians, and $\alpha = 2$. For smaller datasets, the UCI Darwin dataset was used to study Alzheimer’s disease prediction, mapping classes \texttt{P} and \texttt{H} to binary targets, imputing missing values, encoding categorical variables, and applying $\alpha = 2$. The Iris dataset (150 samples, 4 features) was transformed into a binary classification task by grouping Setosa against Versicolor and Virginica, with $\alpha = 1.5$. Finally, the Wine dataset (178 samples, 13 features) was converted into a binary task by grouping one cultivar against the others, with $\alpha = 1.5$. Together, these datasets provided a diverse evaluation environment spanning large-scale, high-dimensional data to smaller, feature-limited cases.

\section{8. Results and Discussion}

We evaluate the FIGRF model on multiple datasets using accuracy, precision, recall, and F1-score, while also examining ranked feature importance scores, usage frequency across decision trees, and confusion matrices for sample-level classification, with a standard RF model applied under the same preprocessing pipeline serving as the baseline. For high-dimensional datasets, the UCI Credit Card Client dataset was optimized through SA to 140 estimators and a maximum depth of 20, achieving 82.1\% accuracy, 70\% precision, 31.76\% recall, and a 43.71\% F1-score as shown in Table~\ref{tab:rf_performance_comparison}, surpassing the baseline in accuracy and precision though slightly lower in recall and F1-score; FS-RL proposed in \cite{bouchlaghem2024novel} achieved lower accuracy, as reported in Table~\ref{tab:rf_performance_comparison_RW}. Feature importance analysis highlighted the current month’s payment status \texttt{PAY\_0}, along with \texttt{credit limit}, \texttt{previous payment}, and \texttt{billing} features, as the most influential predictors. On the IoTID20 dataset, SA converged to 100 estimators and a depth of 18, producing perfect performance across all four metrics, with zero false predictions on over 100,000 samples, while \texttt{Timestamp}, \texttt{Flow\_ID}, and \texttt{Dst\_Port} emerged as dominant contributors in Figure~\ref{imp_iotid20}, outperforming the baseline’s 99\% performance. For the UCI Bank Marketing dataset, SA identified 50 estimators and a depth of 8 as optimal, yielding improved recall (55.72\% vs. 48.77\%) and F1-score (59.91\% vs. 55.81\%) relative to the baseline, while accuracy remained around 91\%, with \texttt{duration}, \texttt{Euribor3m}, and \texttt{cons.conf.idx} most frequently selected. On the WUSTL-IIoT 2021 dataset, FIGRF achieved near-perfect performance (about 99.99\% across all metrics), consistent with the baseline, while features such as \texttt{SIntPktl}, \texttt{sTtl}, and \texttt{DIntPkt} were found most critical. For low-dimensional datasets, the Darwin dataset tuned with parameters (53, 10) achieved flawless performance across all metrics, outperforming the baseline accuracy of 94\%, with high contributions from features like \texttt{ID}, \texttt{air\_time5}, \texttt{air\_time17}, and \texttt{paper\_time22}. The Iris dataset, tuned with parameters (71, 10), also achieved 100\% accuracy, precision, recall, and F1-score, matching the baseline but surpassing the Chi2-PSOGWO method  \cite{abdo2024optimized} in Table~\ref{tab:rf_performance_comparison_RW}, with \texttt{petal width (cm)} and \texttt{petal length (cm)} identified as most significant. Finally, the Wine dataset optimized with (109, 13) reached perfect classification on the test set, consistent with the baseline and the ASOs+SA method \cite{ghosh2020aso}, while Chi2-PSOGWO \cite{abdo2024optimized} slightly underperformed with 99.4\% accuracy; feature analysis emphasized \texttt{proline}, \texttt{flavanoids}, and \texttt{alcohol} as the top discriminators. Overall, the FIGRF framework, integrating probabilistic feature selection with SA-based hyperparameter tuning, consistently matched or outperformed standard RF across both high- and low-dimensional datasets, delivering notable gains in recall and F1-score for imbalanced datasets like IoTID20 \cite{ullah2020scheme} and Bank Marketing, while preserving near-perfect prediction on balanced datasets such as Iris and Wine. These results, supported by Tables~\ref{tab:rf_performance_comparison} and \ref{tab:rf_performance_comparison_RW} and visualized in Figures~\ref{imp_iotid20}--\ref{imp_darwin}, demonstrate that FIGRF achieved zero false predictions in several cases, aligned feature importance with domain relevance, and surpassed competing methodologies, highlighting the effectiveness of combining guided feature selection with metaheuristic optimization for interpretable and high-performing tree models.

\section{CONCLUSION}
This research introduced a Feature-Importance Guided Random Forest framework integrating probabilistic feature-weighted sampling with Simulated Annealing–driven hyperparameter optimization. By coupling relevance-aware feature selection with adaptive search dynamics, the method advances the conventional Random Forest paradigm while preserving model interpretability. Empirical evaluation across heterogeneous benchmark datasets demonstrates its robustness and scalability across multi-dimensional datasets, underscoring its utility as a practical extension of ensemble learning architectures.\vspace*{-8pt}

\def\refname{REFERENCES}

\vspace*{-8pt}

\begin{IEEEbiography}{Kowshik Balasubramanian}{\,}received the
B.E. degree in electronics and
instrumentation engineering from Anna
University, India in 2024. He holds a
design patent on stress detection biosensor
in India. He is pursuing M.Sc. degree in
information technology management and
has been contributing to the Tecore
Networks Laboratory at Florida Atlantic
University, Boca Raton, FL, USA, since December 2024. His
research interests include machine learning, deep learning,
data science, artificial intelligence and the internet of things. 
\end{IEEEbiography}

\begin{IEEEbiography}{Andre Williams}{\,}received a Bachelor of Science in Electrical Engineering from Texas A\&M, a Master of Science from Florida Atlantic University, and is currently working on a Ph.D. from Florida Atlantic University. His research interests include Anomaly Detection, Denoising of Hyperspectral data using transformers and convolutional networks, and using large language models in practical applications.\vspace*{-2pt}
\end{IEEEbiography}

\begin{IEEEbiography}{Ismail Butun} {\,}received
the B.Sc. and M.Sc. degrees in electrical
and electronics engineering from
Hacettepe University, Türkiye, and the
M.Sc. and Ph.D. degrees in electrical
engineering from the University of South
Florida, Tampa, FL, USA, in 2009 and
2013, respectively. From 2014 to 2017, he
was an Assistant Professor with the
Department of Mechatronics Engineering, Bursa Technical
University, and then the Department of Computer Engineering,
Abdullah Gul University. From 2016 to 2021, he was a
Postdoctoral Fellow for various universities, such as the
University of Delaware, Mid Sweden University, and Chalmers
University of Technology. From 2021 and 2023, he was with
the School of Electrical Engineering and Computer Science,
KTH Royal Institute of Technology. Since February 2024, he
has been contributing to the Tecore Networks Laboratory,
Florida Atlantic University, Boca Raton, FL, USA. He has more
than 65 publications in peer-reviewed scientific international
journals and conference proceedings, along with 3500 citations,
an H-index of 24, and an i10-index of 39. His research interests
include computer networks, wireless communications, WSNs,
the IoT, cyber-physical systems, cryptography, network
security, location privacy, and intrusion detection. He is an
Editor of IEEE Access journal, Sensors (MDPI) journal,
Journal of Sensors (Hindawi), and Nature (Springer). He is a
well-recognized academic reviewer by IEEE, ACM, and
Springer, who served more than 50 various scientific journals
and conferences in the review process of more than 250 articles.
\end{IEEEbiography}

\end{document}